\pdfoutput=1

\documentclass[11pt]{article}

\usepackage{acl}

\usepackage{times}
\usepackage{latexsym}
\usepackage{kotex}
\usepackage{todonotes}
\usepackage{arydshln}
\usepackage{tabularx}
\usepackage{booktabs}
\usepackage{graphicx}
\usepackage{amsmath}
\usepackage{makecell}
\usepackage{array}
\usepackage{tabularx}
\usepackage{multirow}
\usepackage{siunitx}
\usepackage{adjustbox}
\usepackage{rotating}
\usepackage[utf8]{inputenc}
\usepackage{caption}
\usepackage{hyperref}

\usepackage[T1]{fontenc}

\usepackage[utf8]{inputenc}

\usepackage{microtype}

\usepackage{inconsolata}

\usepackage{graphicx}

\title{From Ambiguity to Accuracy: The Transformative Effect of Coreference Resolution on Retrieval-Augmented Generation systems}

\author{
  Youngjoon Jang$^{*,1}$, Seongtae Hong$^{*,1}$, Junyoung Son$^{1}$ \\ \textbf{Sungjin Park$^{2}$}, \textbf{Chanjun Park$^{\dagger,1}$}, \textbf{Heuiseok Lim$^{\dagger,1}$} \\
  \\
  $^1$Korea University \\
  \texttt{\{dew1701, ghdchlwls123, s0ny, bcj1210, limhseok\}@korea.ac.kr} \\
  $^2$Naver Corp \\
  \texttt{sungjin.park@navercorp.com}
}

\newcommand\blfootnote[1]{%
  \begingroup
  \renewcommand\thefootnote{}\footnote{#1}%
  \addtocounter{footnote}{-1}%
  \endgroup
}

\begin{document}
\maketitle
\begin{abstract}
Retrieval-Augmented Generation (RAG) has emerged as a crucial framework in natural language processing (NLP), improving factual consistency and reducing hallucinations by integrating external document retrieval with large language models (LLMs). However, the effectiveness of RAG is often hindered by coreferential complexity in retrieved documents, introducing ambiguity that disrupts in-context learning. In this study, we systematically investigate how entity coreference affects both document retrieval and generative performance in RAG-based systems, focusing on retrieval relevance, contextual understanding, and overall response quality. We demonstrate that coreference resolution enhances retrieval effectiveness and improves question-answering (QA) performance. Through comparative analysis of different pooling strategies in retrieval tasks, we find that mean pooling demonstrates superior context capturing ability after applying coreference resolution. In QA tasks, we discover that smaller models benefit more from the disambiguation process, likely due to their limited inherent capacity for handling referential ambiguity. With these findings, this study aims to provide a deeper understanding of the challenges posed by coreferential complexity in RAG, providing guidance for improving retrieval and generation in knowledge-intensive AI applications.
\end{abstract}

\section{Introduction}
\blfootnote{$^*$Equal contribution.}
\blfootnote{$^\dagger$ Corresponding Author}
With the rapid advancement of large language models (LLMs) and information retrieval technologies, Retrieval-Augmented Generation (RAG) has emerged as a fundamental technique widely adopted across various tasks, including knowledge-intensive applications such as question-answering and dialogue systems~\citep{gan2023large, yang2023large}. By integrating retrieval mechanisms with generative language models, RAG enhances factual consistency, improves knowledge recall, and mitigates issues related to hallucination.

\begin{figure}[t]
\begin{center}
\includegraphics[width=1 \linewidth]{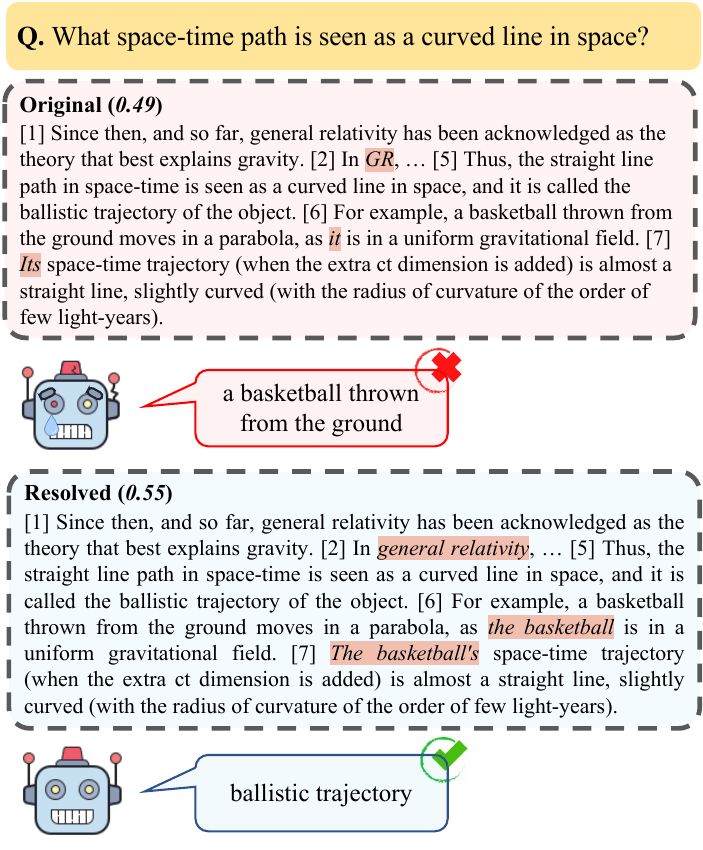}
\end{center}

\caption{Example of changes in similarity and responses resulting from coreference resolution. Similarity scores are indicated in parentheses using NV-Embed-v2, and responses are generated with the Llama-3.2-1B-Instruct model.}
\label{fig:overview}
\end{figure}

Two key challenges in RAG lie in the retrieval of relevant documents from a large corpus and the subsequent in-context learning process, where retrieved documents are leveraged to generate accurate responses. These challenges are particularly pronounced when dealing with documents, as these often contain multiple coreferences to the same entities, making it difficult for language models to resolve coreferential ambiguity effectively~\cite{quoref_coref_reasoning}. In addition, these hinder the ability of LLMs to effectively capture relevant contextual information from the given inputs~\cite{bridging_context_gaps}.



From this perspective, coreferential complexity can hinder a retrieval model’s ability to effectively interpret and represent documents. Specifically, it may prevent the model from accurately capturing the semantic relationships between entities and their references, making it more difficult to align query intentions with the most relevant document. These retrieval errors and drops in relevance propagate throughout the generation process, ultimately reducing the factual accuracy of the responses \cite{llm_canbe_easily_distracted}. Consequently, such accumulated errors undermine user trust in AI-generated answers, weakening confidence in the system’s outputs.

To address these challenges, we aim to systematically investigate the impact of coreferential complexity on each core component of RAG, including document retrieval and in-context learning. 
Through extensive experiments and analysis, our study reveals two key findings: First, In retrieval tasks, models show performance improvements when coreference resolution is applied, with models utilizing mean pooling demonstrating particularly significant gains. This suggests that resolved coreferences enhance the models’ ability to capture document semantics. 
Second, For QA tasks, we find that smaller language models are likely to benefit more from coreference resolution compared to larger models, indicating that coreferential complexity poses a greater challenge for models with limited capacity. 
These findings highlight how coreference resolution can enhance different aspects of RAG systems, with specific benefits depending on the model architecture and task type.





\section{Coreference Resolution} 
Coreference resolution is a technique that identifies and links different expressions referring to the same entity in a text by identifying and replacing them with their explicit forms to eliminate ambiguity~\citep{ng-2010-supervised}. Figure~\ref{fig:overview} illustrates how this technique enhances natural language processing tasks through explicit entity references, using an actual example from the SQuAD2.0 dataset. In the document, ambiguous elements such as abbreviations and pronouns (“GR”, “it”, “Its”) are replaced with their explicit forms (“general relativity”, “the basketball”, “The basketball’s”). Comparing the original and resolved documents, the similarity scores computed by the embedding model show an improvement for the resolved version, demonstrating that coreference resolution effectively enhances the precision of similarity computation for retrieval tasks.
Beyond retrieval performance, coreference resolution significantly impacts question-answering accuracy by strengthening contextual coherence and logical reasoning. The resolved document provides a more traceable reasoning chain, enabling the model to better understand entity relationships and semantics. As demonstrated in our example, the model provides the correct answer with the resolved document while failing with the original document, showing the benefits of this enhanced clarity.
This example clearly illustrates the critical role of coreference resolution in enhancing both document retrieval and question-answering capabilities.

To systematically address coreferential ambiguities, we implement an LLM-powered coreference resolution function $f_\text{coref}$ that transforms ambiguous coreferences into their explicit antecedents. For each document $d_i$, this function produces coreferentially explicit document $d_i^{\prime}$:
\[
d_i^\prime = f_\text{coref}(d_i)
\]
We utilize \textit{gpt-4o-mini}~\citep{hurst2024gpt} to implement this coreference resolution function. The model takes text containing unresolved coreferences as input and produces an output in which multiple expressions referring to the same entity are explicitly linked, maintaining contextual consistency throughout the text. Through this process, we explore how resolving coreferential ambiguity and providing explicit semantic connections in the document impact retrieval and question answering. The detailed prompt design and implementation specifics are described in Section~\ref{sec:exp_setup_prompt}

\section{Experimental Setup}
\sisetup{%
  table-number-alignment = center,
  round-mode = places,
  round-precision = 3
}

\begin{table*}[htbp]
\centering
\small
\setlength{\tabcolsep}{4pt} 
\renewcommand{\arraystretch}{1} 
\resizebox{0.95\textwidth}{!}{
\begin{tabular}{@{}l l l l | *{16}{S[table-format=1.3]} @{}}
\toprule
\multirow{2.3}{*}{\shortstack{Archi\\tecture}} & \multirow{2.3}{*}{Pool} & \multirow{2.3}{*}{Models} & \multirow{2.3}{*}{DocType} & \multicolumn{3}{c}{BELEBELE} & \multicolumn{3}{c}{SQuAD2.0} & \multicolumn{3}{c}{BoolQ} & \multicolumn{3}{c}{NanoSCIDOCS} & \multicolumn{3}{c}{AVG} & {OVR} \\  
\cmidrule(l){5-7}
\cmidrule(l){8-10}
\cmidrule(l){11-13}
\cmidrule(l){14-16}
\cmidrule(l){17-19}
 & & & & {@ 1} & {@ 3} & {@ 5} & {@ 1} & {@ 3} & {@ 5} & {@ 1} & {@ 3} & {@ 5} & {@ 1} & {@ 3} & {@ 5} & {@ 1} & {@ 3} & {@ 5} \\
\midrule

\multirow{9}{*}{\rotatebox{90}{ENCODER}} 
 & \multirow{4}{*}{Mean} 
   & \multirow{2}{*}{e5-large-v2}
   & Original  & 0.922 & \textbf{0.952} & 0.955 & 0.802 & 0.881 & 0.891 & \textbf{0.839} & \textbf{0.905} & \textbf{0.913} & 0.520 & \textbf{0.404} & \textbf{0.359} & 0.809 & 0.812 & \textbf{0.810} & 0.811   \\  
 &    &  
   & \; + \textsc{C}$\cdot$\textsc{R}         & \textbf{0.923} & 0.948 & \textbf{0.956} & \textbf{0.816} & \textbf{0.893} & \textbf{0.902} & 0.833 & 0.902 & 0.911 & \textbf{0.520} & 0.400 & 0.352 & \textbf{0.814} & \textbf{0.813} & 0.809 & \textbf{0.812} \\ 
\cmidrule(l){3-20}
 &    & \multirow{2}{*}{stella\_en\_400M\_v5}
   & Original & 0.910 & 0.946 & 0.949 & 0.767 & \textbf{0.851} & \textbf{0.866} & \textbf{0.838} & 0.907 & 0.915 & 0.480 & \textbf{0.386} & 0.345 & 0.785 & 0.799 & 0.803 & 0.796 \\
 &    &  
   & \; + \textsc{C}$\cdot$\textsc{R}          & \textbf{0.920} & \textbf{0.950} & \textbf{0.954} & 0.767 & 0.849 & 0.864 & 0.837 & \textbf{0.907} & \textbf{0.915} & \textbf{0.500} & 0.384 & \textbf{0.349} & \textbf{0.790} & \textbf{0.799} & \textbf{0.804} & \textbf{0.798} \\
\cmidrule(l){2-20}
 & \multirow{4}{*}{[CLS]} 
   & \multirow{2}{*}{gte-modernbert-base}
   & Original & 0.892 & 0.932 & 0.937 & 0.778 & 0.862 & 0.876 & \textbf{0.831} & \textbf{0.901} & 0.909 & 0.520 & \textbf{0.452} & \textbf{0.410} & 0.793 & 0.809 & \textbf{0.811} & 0.804 \\
 &    &  
   & \; + \textsc{C}$\cdot$\textsc{R}        & \textbf{0.899} & \textbf{0.936} & \textbf{0.940} & \textbf{0.779} & \textbf{0.863} & \textbf{0.876} & 0.829 & 0.900 & \textbf{0.909} & \textbf{0.520} & 0.448 & 0.391 & \textbf{0.794} & \textbf{0.809} & 0.807 & \textbf{0.804} \\
\cmidrule(l){3-20}
 &    & \multirow{2}{*}{bge-large-en-v1.5}
   & Original & 0.903 & 0.932 & 0.939 & \textbf{0.749} & 0.838 & \textbf{0.854} & 0.831 & 0.899 & 0.908 & 0.480 & \textbf{0.395} & \textbf{0.364} & 0.776 & \textbf{0.792} & 0.799 & 0.789 \\
 &    &  
   & \; + \textsc{C}$\cdot$\textsc{R}        & \textbf{0.912} & \textbf{0.938} & \textbf{0.944} & 0.747 & \textbf{0.838} & 0.853 & \textbf{0.833} & \textbf{0.901} & \textbf{0.909} & \textbf{0.480} & 0.382 & 0.359 & \textbf{0.777}	& 0.791 & \textbf{0.800} & \textbf{0.789} \\

\midrule
\midrule

\multirow{9}{*}{\rotatebox{90}{DECODER}}
 & \multirow{4}{*}{Mean} 
   & \multirow{2}{*}{NV-Embed-v2}
   & Original & 0.959 & 0.977 & \textbf{0.978} & 0.865 & 0.927 & 0.933 & 0.874 & 0.935 & 0.941 & 0.460 & 0.405 & \textbf{0.356} & 0.836 & 0.842 & 0.836 & 0.838 \\
 &    &  
   & \; + \textsc{C}$\cdot$\textsc{R}        & \textbf{0.959} & \textbf{0.977} & 0.977 & \textbf{0.873} & \textbf{0.933} & \textbf{0.938} & \textbf{0.874} & \textbf{0.935} & \textbf{0.941} & \textbf{0.480} & \textbf{0.414} & 0.353 & \textbf{0.843} & \textbf{0.845} & \textbf{0.836} & \textbf{0.841}  \\
\cmidrule(l){3-20}
 &    & \multirow{2}{*}{LLM2Vec}
   & Original & 0.938 & 0.964 & 0.967 & 0.835 & 0.904 & 0.913 & 0.854 & 0.922 & 0.929 & 0.440 & 0.408 & 0.358 & 0.814 & 0.827 & 0.824 & 0.822 \\
 &    &  
   & \; + \textsc{C}$\cdot$\textsc{R}        & \textbf{0.941} & \textbf{0.965} & \textbf{0.968} & \textbf{0.839} & \textbf{0.907} & \textbf{0.916} & \textbf{0.854} & \textbf{0.922} & \textbf{0.929} & \textbf{0.500} & \textbf{0.424} & \textbf{0.372} & \textbf{0.826} & \textbf{0.831} & \textbf{0.827} & \textbf{0.828} \\
\cmidrule(l){2-20}
 & \multirow{4}{*}{Last} 
   & \multirow{2}{*}{gte-Qwen2-1.5B}
   & Original & 0.938 & \textbf{0.961} & 0.963 & 0.820 & 0.891 & 0.901 & 0.823 & 0.893 & 0.904 & 0.520 & 0.428 & 0.387 & 0.816 & 0.816 & 0.812 & 0.815 \\
 &    &  
   & \; + \textsc{C}$\cdot$\textsc{R}        & \textbf{0.940} & 0.959 & \textbf{0.964} & \textbf{0.820} & \textbf{0.891} & \textbf{0.901} & \textbf{0.825} & \textbf{0.895} & \textbf{0.906} & \textbf{0.520} & \textbf{0.435} & \textbf{0.392} & \textbf{0.816} & \textbf{0.818} & \textbf{0.814} & \textbf{0.816} \\
\cmidrule(l){3-20}
 &    & \multirow{2}{*}{Linq-Embed-Mistral}
   & Original & \textbf{0.944} & 0.967 & 0.969 & \textbf{0.800} & \textbf{0.885} & \textbf{0.895} & 0.876 & 0.937 & 0.942 & 0.460 & 0.407 & 0.360 & 0.810 & 0.828 & 0.830 & 0.823 \\
 &    &  
   & \; + \textsc{C}$\cdot$\textsc{R}        & 0.942 & \textbf{0.967} & \textbf{0.969} & 0.798 & 0.882 & 0.892 & \textbf{0.877} & \textbf{0.937} & \textbf{0.942} & \textbf{0.500} & \textbf{0.423} & \textbf{0.373} & \textbf{0.815} & \textbf{0.830} & \textbf{0.832} & \textbf{0.826} \\

\bottomrule
\end{tabular}
}
\caption{Performance of retrieval tasks with and without coreference resolution. The @k indicates the top k nDCG results. For each comparison, the higher score is highlighted in \textbf{bold}.}

\label{tab:model_comparison}
\end{table*}

\paragraph{Models}

We evaluate a variety of publicly accessible embedding models with different architectures and pooling methods to evaluate retrieval performance for both the original document and the coreference-resolved document. For encoder-based embedding models, we use \textit{e5-large-v2}~\cite{wang2022text}, \textit{stella\_en\_400M\_v5}~\cite{zhang2025jasperstelladistillationsota}, \textit{bge-large-en-v1.5}~\cite{bge_embedding}, and \textit{gte-modernbert-base}~\cite{zhang2024mgtegeneralizedlongcontexttext}. As decoder-based models, we employ \textit{LLM2Vec-Meta-Llama-3-8B-Instruct-mntp-supervised}~\cite{behnamghader2024llm2veclargelanguagemodels} which we refer to as \textit{LLM2Vec}, \textit{NV-Embed-v2}~\cite{lee2025nvembedimprovedtechniquestraining}, \textit{Linq-Embed-Mistral}~\cite{choi2024linq}, and \textit{gte-Qwen2-1.5B-instruct}~\cite{li2023towards}. 


To evaluate how coreference resolution affects LLMs’ understanding and answer generation capabilities, we conduct experiments with various instruction-tuned models: \textit{Llama3.2-3B-Instruct}, \textit{Llama3.1-8B-Instruct}~\citep{dubey2024llama}, \textit{Qwen2.5-3B-Instruct}, \textit{Qwen2.5-7B-Instruct}~\citep{yang2024qwen2}, \textit{gemma-2-2b-it}, \textit{gemma-2-9b-it}~\citep{team2024gemma}, \textit{Mistral-7B-Instruct-v0.3}~\citep{jiang2023mistral}.

\paragraph{Datasets} 
To evaluate the effect of coreferential complexity in retrieval performance, we conduct experiments on four datasets: BELEBELE~\citep{bandarkar2023belebele}, which is designed for Machine Reading Comprehension (MRC) tasks, SQuAD2.0~\citep{rajpurkar2018know}, a QA dataset based on Wikipedia, BoolQ~\citep{clark2019boolq}, designed for yes/no questions, and NanoSCIDOCS~\citep{cohan2020specter}, which is a subset of SCIDOCS dataset, specifically designed for retrieval tasks. For the QA datasets, we adapt the question-document pairs for retrieval evaluation. Details about data preprocessing and extra experiment details can be found in Appendix~\ref{sec:exp_setup}.



\paragraph{Metrics}
We use nDCG@k(k=1,3,5) to evaluate retrieval performance. nDCG evaluates retrieval ranking quality by measuring both relevance and position of results with logarithmic position discount.
For evaluating QA performance, we calculate the log likelihood on benchmarks such as the BELEBELE and BoolQ datasets for accuracy measurement, and use the F1-score for SQuAD2.0. All experiments are conducted using the library\footnote{\scriptsize\url{https://github.com/EleutherAI/lm-evaluation-harness}} to ensure replicability.

\section{Experimental Results and Analysis}
\subsection{Impact of Coreference Resolution on Retrieval Performance}

Table~\ref{tab:model_comparison} presents a comparison of retrieval performance between original documents and their coreference-resolved versions across different embedding models. Our experiments demonstrate that addressing coreference issues consistently improves retrieval performance across all evaluation metrics, likely due to more explicit and traceable entity references in document representations.
The performance improvement is particularly pronounced in decoder-based models, with \textit{LLM2Vec} shows the most significant gains in the average score, improving by 0.012, 0.004, and 0.003 points for nDCG@k (k=1, 3, 5), respectively. 
These results demonstrate that coreference resolution enhances the overall performance of retrieval tasks, particularly in decoder-based embedding models.



Furthermore, we observe a trend along with the choice of pooling strategies in embedding models. Specifically, models employing mean pooling (e.g., \textit{e5-large-v2}, \textit{stella\_en\_400M\_v5}, \textit{NV-Embed-v2}, and \textit{LLM2Vec}) exhibit a more clear performance gain from coreference resolution compared to models utilizing [CLS] token or last token pooling. 
This phenomenon can be explained by mean pooling’s equal treatment of all tokens. By replacing pronouns with their actual antecedents, more meaningful semantic representations are captured, as each token now carries more explicit semantic information rather than abstract references.
This observation aligns with previous research suggesting that mean pooling is particularly useful for capturing the overall semantics of text data~\citep{zhao2022augment}. While [CLS] token and last token pooling methods also show improvements with coreference resolution, their reliance on a single-token representation for the entire document embedding leads to relatively smaller gains compared to mean pooling. As shown in Table~\ref{tab:referential_complexity}, coreference resolution tends to increase document length by replacing pronouns with their antecedents. This characteristic further amplifies the advantage of mean pooling, which can more effectively integrate information across varying text lengths.
These findings highlight the synergistic relationship between mean pooling and coreference resolution in enhancing document representation.


\subsection{Impact of Coreference Resolution on Question Answering Performance}
Table~\ref{tab:qa} examines the impact of coreference resolution on QA tasks across different model architectures and sizes. We observe consistent performance improvements across all models and tasks, aligning with previous findings on the benefits of coreference resolution in question answering~\citep{bridging_context_gaps}.

\begin{table}[tp]
\centering
\renewcommand{\arraystretch}{1} 
\resizebox{\columnwidth}{!}{%
\begin{tabular}{%
  >{\arraybackslash}p{2.0cm}   
  >{\arraybackslash}p{1.7cm}   
  >{\centering\arraybackslash}p{1.9cm}   
  >{\centering\arraybackslash}p{1.9cm}   
  >{\centering\arraybackslash}p{1.9cm}   
}
\toprule
\textbf{Models} & \textbf{DocType} & \textbf{BoolQ} & \textbf{BELEBELE} & \textbf{SQuAD}\\
\midrule
\multirow{2}{*}{\makecell[l]{Llama3.2-3B\\-Instruct}}
   & Orginal & 0.7636 & 0.8122 & 0.6437 \\
   & \; + \textsc{C}$\cdot$\textsc{R}       & \textbf{0.7642} & \textbf{0.8389} & \textbf{0.6888} \\
\midrule
\multirow{2}{*}{\makecell[l]{Llama-3.1-8B\\-Instruct}}
   & Orginal & 0.8202 & 0.8833 & 0.5583 \\
   & \; + \textsc{C}$\cdot$\textsc{R}       & \textbf{0.8205} & \textbf{0.9133} & \textbf{0.7827} \\
\midrule
\multirow{2}{*}{\makecell[l]{Qwen2.5-3B\\-Instruct}}
   & Orginal & 0.7801 & 0.7800 & 0.2972 \\
   & \; + \textsc{C}$\cdot$\textsc{R}       & \textbf{0.7804} & \textbf{0.8578} & \textbf{0.5500} \\
\midrule
\multirow{2}{*}{\makecell[l]{Qwen2.5-7B\\-Instruct}}
   & Orginal & 0.8599 & 0.8622 & 0.3980 \\
   & \; + \textsc{C}$\cdot$\textsc{R}       & 0.8599 & \textbf{0.9022} & \textbf{0.7977} \\
\midrule
\multirow{2}{*}{\makecell[l]{gemma-2\\-2b-it}}
   & Orginal & 0.8006 & 0.2633 & 0.5185 \\
   & \; + \textsc{C}$\cdot$\textsc{R}       & \textbf{0.8015} & \textbf{0.3067} & \textbf{0.6209} \\
\midrule
\multirow{2}{*}{\makecell[l]{gemma-2\\-9b-it}}
   & Orginal & 0.8645 & 0.5411 & 0.7646 \\
   & \; + \textsc{C}$\cdot$\textsc{R}       & \textbf{0.8651} & \textbf{0.5467} & \textbf{0.8423} \\
\midrule
\multirow{2}{*}{\makecell[l]{Mistral-7B\\-Instruct-v0.3}}
   & Orginal & 0.8321 & 0.8500 & 0.4080 \\
   & \; + \textsc{C}$\cdot$\textsc{R}       & \textbf{0.8349} & \textbf{0.8511} & \textbf{0.4396} \\
\bottomrule
\end{tabular}
}
\caption{Performance of QA tasks on coreference resolution. The higher score is highlighted in bold.}
\label{tab:qa}
\end{table}

Notably, smaller models tend to achieve greater performance gains through coreference resolution compared to their larger variants. For instance, in BoolQ, \textit{Qwen2.5-3B-Instruct} shows an improvement of 0.0003 compared to no improvement in the 7B version, and \textit{gemma-2-2b-it} improves by 0.0009 whereas the \textit{9b} model shows an improvement of 0.0006. This pattern becomes more pronounced in the Belebele task, where \textit{Qwen2.5-3B-Instruct} demonstrates an improvement of 0.0778, substantially higher than the 0.0400 gain of its \textit{7B} variant, and \textit{gemma-2-2b-it} achieves a 0.0434 improvement compared to the minimal 0.0056 gain in the \textit{9b} version. As Table~\ref{tab:referential_complexity} shows, applying coreference resolution reduces the number of pronouns, thereby decreasing coreferential complexity. This more explicit representation facilitates easier contextual understanding, particularly benefiting smaller language models.

Interestingly, we find that in SQuAD2.0, some small models with given coreference-resolved document perform comparably to or even surpass larger models using original document. For example, \textit{gemma-2-2b-it} and \textit{Qwen2.5-3B-Instruct} achieve F1-scores of 0.6209 and 0.5500 respectively with coreference-resolved document, which are similar to or higher than the baseline performance of larger models such as \textit{Llama3.1-8B-Instruct}, \textit{Qwen2.5-7B-Instruct}, and \textit{Mistral-7B-Instruct-v0.3} (scoring 0.5583, 0.3980, and 0.4080 respectively). These findings collectively suggest that coreference resolution is impactful for QA tasks, where reducing coreferential complexity directly aids models by facilitating improved contextual understanding.

\section{Conclusion}
This study investigates the effectiveness of coreference resolution in enhancing natural language understanding across retrieval and question answering tasks. Our comprehensive analysis reveals several key findings. First, dense embedding models show consistent improvements in retrieval performance when coreference resolution is applied, with mean pooling strategies particularly benefiting from more explicit entity representations. Second, the impact of coreference resolution varies across model architectures and sizes: while it enhances performance across all scales, smaller language models show particularly notable improvements, sometimes achieving comparable performance to larger models when given coreference-resolved document. These findings highlight how reducing coreferential complexity can effectively enhance model performance, contributing to our understanding of how to improve contextual comprehension in language models. Our work provides valuable insights for future research in optimizing both retrieval systems and question answering models through better handling of coreferential relationships.

\section*{Limitations}
Despite the contributions of this study, there are several limitations that should be acknowledged. We identify potential biases arising from the use of GPT-4o-mini for coreference resolution, as the model’s interpretations may not always align with human understanding, leading to possible discrepancies. Additionally, despite employing diverse datasets (e.g., BELEBELE, SQuAD2.0, BoolQ, NanoSCIDOCS), our approach may not fully capture the complexities of specialized or highly technical text, indicating the need for broader, domain-specific evaluation. Finally, while providing explicit references can increase clarity by grounding model outputs, this method can sometimes constrain the generative flexibility of language models, thereby limiting their ability to produce a wide range of natural-sounding responses. Balancing clarity with generative versatility thus remains a critical direction for future research.

\section*{Ethics Statement}
This study acknowledges several ethical considerations. The coreference resolution process may unintentionally perpetuate or amplify existing biases, particularly in sensitive areas such as gender or cultural references, necessitating regular audits of training data. We have documented potential biases and limitations in the use of GPT-4o-mini throughout our research. This paper involved the use of GPT-4o for supporting aspects of the manuscript preparation, such as improving clarity and grammar, while all intellectual contributions, experimental designs, analyses, and core findings remain the responsibility of the authors. Additionally, we acknowledge that the computational cost of coreference resolution raises environmental concerns, and its application in critical decision-making processes requires careful consideration. We maintain transparency in our methodologies to facilitate reproducibility and further research in this area.

\section*{Acknowledgments}
This work was supported by ICT Creative Consilience Program through the Institute of Information \& Communications Technology Planning \& Evaluation(IITP) grant funded by the Korea government(MSIT) (IITP-2025-RS-2020-II201819) and Institute of Information \& communications Technology Planning \& Evaluation(IITP) under the Leading Generative AI Human Resources Development(IITP-2025-R2408111) grant funded by the Korea government(MSIT) and Institute for Information \& communications Technology Promotion(IITP) grant funded by the Korea government(MSIT) (RS-2024-00398115, Research on the reliability and coherence of outcomes produced by Generative AI)


\bibliography{custom}

@article{bridging_context_gaps,
  title={Bridging context gaps: Leveraging coreference resolution for long contextual understanding},
  author={Liu, Yanming and Peng, Xinyue and Cao, Jiannan and Bo, Shi and Shen, Yanxin and Zhang, Xuhong and Cheng, Sheng and Wang, Xun and Yin, Jianwei and Du, Tianyu},
  journal={arXiv preprint arXiv:2410.01671},
  year={2024}
}

@article{quoref_coref_reasoning,
  title={Quoref: A reading comprehension dataset with questions requiring coreferential reasoning},
  author={Dasigi, Pradeep and Liu, Nelson F and Marasovi{\'c}, Ana and Smith, Noah A and Gardner, Matt},
  journal={arXiv preprint arXiv:1908.05803},
  year={2019}
}

@inproceedings{llm_canbe_easily_distracted,
  title={Large language models can be easily distracted by irrelevant context},
  author={Shi, Freda and Chen, Xinyun and Misra, Kanishka and Scales, Nathan and Dohan, David and Chi, Ed H and Sch{\"a}rli, Nathanael and Zhou, Denny},
  booktitle={International Conference on Machine Learning},
  pages={31210--31227},
  year={2023},
  organization={PMLR}
}

@inproceedings{gan2023large,
  title={Large language models in education: Vision and opportunities},
  author={Gan, Wensheng and Qi, Zhenlian and Wu, Jiayang and Lin, Jerry Chun-Wei},
  booktitle={2023 IEEE international conference on big data (BigData)},
  pages={4776--4785},
  year={2023},
  organization={IEEE}
}

@article{yang2023large,
  title={Large language models in health care: Development, applications, and challenges},
  author={Yang, Rui and Tan, Ting Fang and Lu, Wei and Thirunavukarasu, Arun James and Ting, Daniel Shu Wei and Liu, Nan},
  journal={Health Care Science},
  volume={2},
  number={4},
  pages={255--263},
  year={2023},
  publisher={Wiley Online Library}
}

@article{wang2022text,
  title={Text Embeddings by Weakly-Supervised Contrastive Pre-training},
  author={Wang, Liang and Yang, Nan and Huang, Xiaolong and Jiao, Binxing and Yang, Linjun and Jiang, Daxin and Majumder, Rangan and Wei, Furu},
  journal={arXiv preprint arXiv:2212.03533},
  year={2022}
}

@misc{zhang2025jasperstelladistillationsota,
      title={Jasper and Stella: distillation of SOTA embedding models}, 
      author={Dun Zhang and Jiacheng Li and Ziyang Zeng and Fulong Wang},
      year={2025},
      eprint={2412.19048},
      archivePrefix={arXiv},
      primaryClass={cs.IR},
      url={https://arxiv.org/abs/2412.19048}, 
}

@misc{bge_embedding,
      title={C-Pack: Packaged Resources To Advance General Chinese Embedding}, 
      author={Shitao Xiao and Zheng Liu and Peitian Zhang and Niklas Muennighoff},
      year={2023},
      eprint={2309.07597},
      archivePrefix={arXiv},
      primaryClass={cs.CL}
}

@article{choi2024linq,
  title={Linq-Embed-Mistral Technical Report},
  author={Choi, Chanyeol and Kim, Junseong and Lee, Seolhwa and Kwon, Jihoon and Gu, Sangmo and Kim, Yejin and Cho, Minkyung and Sohn, Jy-yong},
  journal={arXiv preprint arXiv:2412.03223},
  year={2024}
}

@article{li2023towards,
  title={Towards general text embeddings with multi-stage contrastive learning},
  author={Li, Zehan and Zhang, Xin and Zhang, Yanzhao and Long, Dingkun and Xie, Pengjun and Zhang, Meishan},
  journal={arXiv preprint arXiv:2308.03281},
  year={2023}
}

@misc{behnamghader2024llm2veclargelanguagemodels,
      title={LLM2Vec: Large Language Models Are Secretly Powerful Text Encoders}, 
      author={Parishad BehnamGhader and Vaibhav Adlakha and Marius Mosbach and Dzmitry Bahdanau and Nicolas Chapados and Siva Reddy},
      year={2024},
      eprint={2404.05961},
      archivePrefix={arXiv},
      primaryClass={cs.CL},
      url={https://arxiv.org/abs/2404.05961}, 
}

@misc{lee2025nvembedimprovedtechniquestraining,
      title={NV-Embed: Improved Techniques for Training LLMs as Generalist Embedding Models}, 
      author={Chankyu Lee and Rajarshi Roy and Mengyao Xu and Jonathan Raiman and Mohammad Shoeybi and Bryan Catanzaro and Wei Ping},
      year={2025},
      eprint={2405.17428},
      archivePrefix={arXiv},
      primaryClass={cs.CL},
      url={https://arxiv.org/abs/2405.17428}, 
}

@misc{zhang2024mgtegeneralizedlongcontexttext,
      title={mGTE: Generalized Long-Context Text Representation and Reranking Models for Multilingual Text Retrieval}, 
      author={Xin Zhang and Yanzhao Zhang and Dingkun Long and Wen Xie and Ziqi Dai and Jialong Tang and Huan Lin and Baosong Yang and Pengjun Xie and Fei Huang and Meishan Zhang and Wenjie Li and Min Zhang},
      year={2024},
      eprint={2407.19669},
      archivePrefix={arXiv},
      primaryClass={cs.CL},
      url={https://arxiv.org/abs/2407.19669}, 
}

@article{caramazza1977comprehension,
  title={Comprehension of anaphoric pronouns},
  author={Caramazza, Alfonso and Grober, Ellen and Garvey, Catherine and Yates, Jack},
  journal={Journal of verbal learning and verbal behavior},
  volume={16},
  number={5},
  pages={601--609},
  year={1977},
  publisher={Elsevier}
}

@inproceedings{kantor-globerson-2019-coreference,
    title = "Coreference Resolution with Entity Equalization",
    author = "Kantor, Ben  and
      Globerson, Amir",
    editor = "Korhonen, Anna  and
      Traum, David  and
      M{\`a}rquez, Llu{\'i}s",
    booktitle = "Proceedings of the 57th Annual Meeting of the Association for Computational Linguistics",
    month = jul,
    year = "2019",
    address = "Florence, Italy",
    publisher = "Association for Computational Linguistics",
    url = "https://aclanthology.org/P19-1066/",
    doi = "10.18653/v1/P19-1066",
    pages = "673--677"
}

@article{desmet2003disambiguation,
  title={Disambiguation preferences and corpus frequencies in noun phrase conjunction},
  author={Desmet, Timothy and Gibson, Edward},
  journal={Journal of Memory and Language},
  volume={49},
  number={3},
  pages={353--374},
  year={2003},
  publisher={Elsevier}
}

@book{mitkov1999anaphora,
  title={Anaphora resolution: the state of the art},
  author={Mitkov, Ruslan},
  year={1999},
  publisher={School of Languages and European Studies, University of Wolverhampton~…}
}

@inproceedings{lee-etal-2017-end,
    title = "End-to-end Neural Coreference Resolution",
    author = "Lee, Kenton  and
      He, Luheng  and
      Lewis, Mike  and
      Zettlemoyer, Luke",
    editor = "Palmer, Martha  and
      Hwa, Rebecca  and
      Riedel, Sebastian",
    booktitle = "Proceedings of the 2017 Conference on Empirical Methods in Natural Language Processing",
    month = sep,
    year = "2017",
    address = "Copenhagen, Denmark",
    publisher = "Association for Computational Linguistics",
    url = "https://aclanthology.org/D17-1018/",
    doi = "10.18653/v1/D17-1018",
    pages = "188--197"
}

@article{manning2020emergent,
  title={Emergent linguistic structure in artificial neural networks trained by self-supervision},
  author={Manning, Christopher D and Clark, Kevin and Hewitt, John and Khandelwal, Urvashi and Levy, Omer},
  journal={Proceedings of the National Academy of Sciences},
  volume={117},
  number={48},
  pages={30046--30054},
  year={2020},
  publisher={National Acad Sciences}
}

@inproceedings{blevins-etal-2023-prompting,
    title = "Prompting Language Models for Linguistic Structure",
    author = "Blevins, Terra  and
      Gonen, Hila  and
      Zettlemoyer, Luke",
    editor = "Rogers, Anna  and
      Boyd-Graber, Jordan  and
      Okazaki, Naoaki",
    booktitle = "Proceedings of the 61st Annual Meeting of the Association for Computational Linguistics (Volume 1: Long Papers)",
    month = jul,
    year = "2023",
    address = "Toronto, Canada",
    publisher = "Association for Computational Linguistics",
    url = "https://aclanthology.org/2023.acl-long.367/",
    doi = "10.18653/v1/2023.acl-long.367",
    pages = "6649--6663"
}

@inproceedings{lelanguage,
  title={Are Language Models Robust Coreference Resolvers?},
  author={Le, Nghia T and Ritter, Alan},
  booktitle={First Conference on Language Modeling}
}

@inproceedings{gan2024assessing,
  title={Assessing the Capabilities of Large Language Models in Coreference: An Evaluation},
  author={Gan, Yujian and Poesio, Massimo and Yu, Juntao},
  booktitle={Proceedings of the 2024 Joint International Conference on Computational Linguistics, Language Resources and Evaluation (LREC-COLING 2024)},
  pages={1645--1665},
  year={2024}
}

@inproceedings{chen-etal-2024-dense,
    title = "Dense {X} Retrieval: What Retrieval Granularity Should We Use?",
    author = "Chen, Tong  and
      Wang, Hongwei  and
      Chen, Sihao  and
      Yu, Wenhao  and
      Ma, Kaixin  and
      Zhao, Xinran  and
      Zhang, Hongming  and
      Yu, Dong",
    editor = "Al-Onaizan, Yaser  and
      Bansal, Mohit  and
      Chen, Yun-Nung",
    booktitle = "Proceedings of the 2024 Conference on Empirical Methods in Natural Language Processing",
    month = nov,
    year = "2024",
    address = "Miami, Florida, USA",
    publisher = "Association for Computational Linguistics",
    url = "https://aclanthology.org/2024.emnlp-main.845/",
    doi = "10.18653/v1/2024.emnlp-main.845",
    pages = "15159--15177"
}

@inproceedings{chai-etal-2022-evaluating,
    title = "Evaluating Coreference Resolvers on Community-based Question Answering: From Rule-based to State of the Art",
    author = "Chai, Haixia  and
      Moosavi, Nafise Sadat  and
      Gurevych, Iryna  and
      Strube, Michael",
    editor = "Ogrodniczuk, Maciej  and
      Pradhan, Sameer  and
      Nedoluzhko, Anna  and
      Ng, Vincent  and
      Poesio, Massimo",
    booktitle = "Proceedings of the Fifth Workshop on Computational Models of Reference, Anaphora and Coreference",
    month = oct,
    year = "2022",
    address = "Gyeongju, Republic of Korea",
    publisher = "Association for Computational Linguistics",
    url = "https://aclanthology.org/2022.crac-1.7/",
    pages = "61--73"
}

@inproceedings{wu-etal-2021-coreference,
    title = "Coreference Reasoning in Machine Reading Comprehension",
    author = "Wu, Mingzhu  and
      Moosavi, Nafise Sadat  and
      Roth, Dan  and
      Gurevych, Iryna",
    editor = "Zong, Chengqing  and
      Xia, Fei  and
      Li, Wenjie  and
      Navigli, Roberto",
    booktitle = "Proceedings of the 59th Annual Meeting of the Association for Computational Linguistics and the 11th International Joint Conference on Natural Language Processing (Volume 1: Long Papers)",
    month = aug,
    year = "2021",
    address = "Online",
    publisher = "Association for Computational Linguistics",
    url = "https://aclanthology.org/2021.acl-long.448/",
    doi = "10.18653/v1/2021.acl-long.448",
    pages = "5768--5781"
}

@article{zhao2022augment,
  title={Augment BERT with average pooling layer for Chinese summary generation},
  author={Zhao, Shuai and You, Fucheng and Chang, Wen and Zhang, Tianyu and Hu, Man},
  journal={Journal of Intelligent \& Fuzzy Systems},
  volume={42},
  number={3},
  pages={1859--1868},
  year={2022},
  publisher={IOS Press}
}

@article{hurst2024gpt,
  title={Gpt-4o system card},
  author={Hurst, Aaron and Lerer, Adam and Goucher, Adam P and Perelman, Adam and Ramesh, Aditya and Clark, Aidan and Ostrow, AJ and Welihinda, Akila and Hayes, Alan and Radford, Alec and others},
  journal={arXiv preprint arXiv:2410.21276},
  year={2024}
}

@unpublished{spacy2,
    AUTHOR = {Honnibal, Matthew and Montani, Ines},
    TITLE  = {{spaCy 2}: Natural language understanding with {B}loom embeddings, convolutional neural networks and incremental parsing},
    YEAR   = {2017},
    Note   = {To appear}
}

@article{clark2019boolq,
  title={BoolQ: Exploring the surprising difficulty of natural yes/no questions},
  author={Clark, Christopher and Lee, Kenton and Chang, Ming-Wei and Kwiatkowski, Tom and Collins, Michael and Toutanova, Kristina},
  journal={arXiv preprint arXiv:1905.10044},
  year={2019}
}

@article{cohan2020specter,
  title={Specter: Document-level representation learning using citation-informed transformers},
  author={Cohan, Arman and Feldman, Sergey and Beltagy, Iz and Downey, Doug and Weld, Daniel S},
  journal={arXiv preprint arXiv:2004.07180},
  year={2020}
}

@article{rajpurkar2018know,
  title={Know what you don't know: Unanswerable questions for SQuAD},
  author={Rajpurkar, Pranav and Jia, Robin and Liang, Percy},
  journal={arXiv preprint arXiv:1806.03822},
  year={2018}
}

@article{bandarkar2023belebele,
  title={The belebele benchmark: a parallel reading comprehension dataset in 122 language variants},
  author={Bandarkar, Lucas and Liang, Davis and Muller, Benjamin and Artetxe, Mikel and Shukla, Satya Narayan and Husa, Donald and Goyal, Naman and Krishnan, Abhinandan and Zettlemoyer, Luke and Khabsa, Madian},
  journal={arXiv preprint arXiv:2308.16884},
  year={2023}
}

@article{dubey2024llama,
  title={The llama 3 herd of models},
  author={Dubey, Abhimanyu and Jauhri, Abhinav and Pandey, Abhinav and Kadian, Abhishek and Al-Dahle, Ahmad and Letman, Aiesha and Mathur, Akhil and Schelten, Alan and Yang, Amy and Fan, Angela and others},
  journal={arXiv preprint arXiv:2407.21783},
  year={2024}
}

@article{team2024gemma,
  title={Gemma 2: Improving open language models at a practical size},
  author={Team, Gemma and Riviere, Morgane and Pathak, Shreya and Sessa, Pier Giuseppe and Hardin, Cassidy and Bhupatiraju, Surya and Hussenot, L{\'e}onard and Mesnard, Thomas and Shahriari, Bobak and Ram{\'e}, Alexandre and others},
  journal={arXiv preprint arXiv:2408.00118},
  year={2024}
}

@article{yang2024qwen2,
  title={Qwen2. 5 technical report},
  author={Yang, An and Yang, Baosong and Zhang, Beichen and Hui, Binyuan and Zheng, Bo and Yu, Bowen and Li, Chengyuan and Liu, Dayiheng and Huang, Fei and Wei, Haoran and others},
  journal={arXiv preprint arXiv:2412.15115},
  year={2024}
}

@article{jiang2023mistral,
  title={Mistral 7B},
  author={Jiang, Albert Q and Sablayrolles, Alexandre and Mensch, Arthur and Bamford, Chris and Chaplot, Devendra Singh and Casas, Diego de las and Bressand, Florian and Lengyel, Gianna and Lample, Guillaume and Saulnier, Lucile and others},
  journal={arXiv preprint arXiv:2310.06825},
  year={2023}
}

@inproceedings{ng-2010-supervised,
    title = "Supervised Noun Phrase Coreference Research: The First Fifteen Years",
    author = "Ng, Vincent",
    editor = "Haji{\v{c}}, Jan  and
      Carberry, Sandra  and
      Clark, Stephen  and
      Nivre, Joakim",
    booktitle = "Proceedings of the 48th Annual Meeting of the Association for Computational Linguistics",
    month = jul,
    year = "2010",
    address = "Uppsala, Sweden",
    publisher = "Association for Computational Linguistics",
    url = "https://aclanthology.org/P10-1142/",
    pages = "1396--1411"
}

\appendix

\clearpage

\section{Related Work}

\subsection{Coreference Resolution}
Coreference Resolution plays a crucial role in understanding and representing text. Previous studies have demonstrated that accurately identifying and linking expressions referring to the same entity within a text serves as a fundamental component of natural language understanding~\citep{caramazza1977comprehension, kantor-globerson-2019-coreference, desmet2003disambiguation}. In particular, coreference resolution is considered one of the complex tasks that requires not only grammatical agreement but also semantic coherence and understanding of discourse structure ~\citep{mitkov1999anaphora}. 

For Coreference Resolution, \citet{lee-etal-2017-end} first proposed an end-to-end approach that learns the antecedent distribution of all spans in a document, while \citet{manning2020emergent} utilized attention mechanisms to analyze how language models perform coreference resolution. Recent research explores the use of prompts with LLMs for coreference resolution, demonstrating that prompt-based methods can effectively leverage the model’s inherent linguistic knowledge for this task ~\citep{lelanguage, blevins-etal-2023-prompting, gan2024assessing}.

\subsection{Applications in Downstream Tasks}
There have been various attempts to reduce coreferential complexity to downstream tasks. \citet{chen-etal-2024-dense} proposed propositions as self-contained factual units that reduce context dependency caused by coreference in retrieval tasks. Meanwhile,~\citet{wu-etal-2021-coreference},~\citet{chai-etal-2022-evaluating},~\citet{bridging_context_gaps} have shown that coreference resolution techniques can improve long context understanding and answering performance in QA tasks.

In our paper, we evaluate the impact of coreference resolution through prompting in LLMs on both retrieval and QA tasks. Our analysis of dense embedding models shows that coreference resolution consistently improves retrieval performance, with models using mean pooling strategies demonstrating particularly notable gains. For QA tasks, experiments across BoolQ, Belebele, and SQuAD2.0 reveal that while coreference resolution generally enhances performance across all model sizes, smaller language models tend to achieve greater relative improvements compared to their larger variants.

\section{Additional Experiment}
\label{sec:additional_setup}
Since using GPT-4o-mini is relatively expensive, we perform coreference resolution with a small Language Model, Qwen2.5-7B-Instruct~\citep{yang2024qwen2}, and report the retrieval performance of Embedding models and the QA performance of LLMs.

\begin{table}[h!]
\centering
\renewcommand{\arraystretch}{1} 
\resizebox{\columnwidth}{!}{%
\begin{tabular}{%
  >{\arraybackslash}p{2.0cm}   
  >{\arraybackslash}p{1.7cm}   
  >{\centering\arraybackslash}p{1.9cm}   
  >{\centering\arraybackslash}p{1.9cm}   
  >{\centering\arraybackslash}p{1.9cm}   
}
\toprule
\textbf{Models} & \textbf{DocType} & \textbf{BoolQ} & \textbf{BELEBELE} & \textbf{SQuAD}\\
\midrule
\multirow{3}{*}{\makecell[l]{Qwen2.5-3B\\-Instruct}}
   & Orginal & 0.7801 & 0.7800 & 0.2972 \\
   & \textsc{C}$\cdot$\textsc{R}$\cdot$\textsc{Qwen} & 0.7777 & 0.8489 & 0.3023 \\
   & \textsc{C}$\cdot$\textsc{R}       & \textbf{0.7804} & \textbf{0.8578} & \textbf{0.5500} \\
\midrule
\multirow{3}{*}{\makecell[l]{gemma-2\\-2b-it}}
   & Orginal & 0.8006 & 0.2633 & 0.5185 \\
   & \textsc{C}$\cdot$\textsc{R}$\cdot$\textsc{Qwen} & 0.8003 & 0.3044 & \textbf{0.6215} \\
   & \textsc{C}$\cdot$\textsc{R}       & \textbf{0.8015} & \textbf{0.3067} & 0.6209 \\
\midrule
\multirow{3}{*}{\makecell[l]{Mistral-7B\\-Instruct-v0.3}}
   & Orginal & 0.8321 & 0.8500 & 0.4080 \\
   & \textsc{C}$\cdot$\textsc{R}$\cdot$\textsc{Qwen} & 0.8336 & 0.8500 & 0.5742 \\
   & \textsc{C}$\cdot$\textsc{R}       & \textbf{0.8349} & \textbf{0.8511} & \textbf{0.7396} \\
\bottomrule
\end{tabular}
}
\caption{Performance of QA tasks on coreference resolution via Qwen2.5-7B-Instruct. The higher score is highlighted in bold.}
\label{tab:qwen_cr_qa}
\end{table}

\paragraph{QA Performance}
Table~\ref{tab:qwen_cr_qa} shows results for QA tasks on coreference resolution done by Qwen2.5-7B-Instruct. It shows that resolving coreferential complexity by Qwen2.5-7B-Instruct also marginally improves QA performance above all three models.

\paragraph{Retrieval Performance}
As shown in Table~\ref{tab:qwen_retrieval}, results show that using a lightweight model for coreference resolution also improves retrieval performance. 
Particularly, models using mean pooling strategy demonstrates superior performance, which aligns the prior results in our paper.

These results show that resolving coreferential complexity with relatively small and cost-effective models can also improve retrieval performance (especially models utilizing mean pooling) and QA performance.



\begin{table*}[]
\centering
\small
\setlength{\tabcolsep}{4pt} 
\renewcommand{\arraystretch}{1} 
\resizebox{0.95\textwidth}{!}{
\begin{tabular}{@{}l l l l | *{16}{S[table-format=1.3]} @{}}
\toprule
\multirow{2.3}{*}{\shortstack{Archi\\tecture}} & \multirow{2.3}{*}{Pool} & \multirow{2.3}{*}{Models} & \multirow{2.3}{*}{DocType} & \multicolumn{3}{c}{BELEBELE} & \multicolumn{3}{c}{SQuAD2.0} & \multicolumn{3}{c}{BoolQ} & \multicolumn{3}{c}{NanoSCIDOCS} & \multicolumn{3}{c}{AVG} & {OVR} \\  
\cmidrule(l){5-7}
\cmidrule(l){8-10}
\cmidrule(l){11-13}
\cmidrule(l){14-16}
\cmidrule(l){17-19}
 & & & & {@ 1} & {@ 3} & {@ 5} & {@ 1} & {@ 3} & {@ 5} & {@ 1} & {@ 3} & {@ 5} & {@ 1} & {@ 3} & {@ 5} & {@ 1} & {@ 3} & {@ 5} \\
\midrule

\multirow{7}{*}{\rotatebox{90}{ENCODER}} 
 & \multirow{3}{*}{Mean} 
   & \multirow{3}{*}{stella\_en\_400M\_v5}
   & \; Original & 0.910 & 0.946 & 0.949 & 0.767 & 0.851 & 0.866 & 0.838 & 0.907 & 0.915 & 0.480 & 0.386 & 0.345 & 0.785 & 0.799 & 0.803 & 0.796 \\
 &    &  
   & \; \textsc{C}$\cdot$\textsc{R}          & 0.920 & \textbf{0.950} & \textbf{0.954} & 0.767 & 0.849 & 0.864 & 0.837 & 0.907 & 0.915 & 0.500 & 0.384 & 0.349 & 0.790 & 0.799 & 0.804 & 0.798 \\
&    &  
   & \; \textsc{C}$\cdot$\textsc{R}$\cdot$Qwen          & \textbf{0.921} & \textbf{0.950} & \textbf{0.954} & \textbf{0.784} & \textbf{0.865} & \textbf{0.879} & \textbf{0.841} & \textbf{0.910} & \textbf{0.917} & \textbf{0.540} & \textbf{0.438} & \textbf{0.405} & \textbf{0.805} & \textbf{0.814} & \textbf{0.818} & \textbf{0.812} \\
\cmidrule(l){2-20}
 & \multirow{3}{*}{[CLS]} 
 & \multirow{3}{*}{bge-large-en-v1.5}
   & \; Original & 0.903 & 0.932 & 0.939 & \textbf{0.749} & \textbf{0.838} & \textbf{0.854} & 0.831 & 0.899 & 0.908 & \textbf{0.480} & \textbf{0.395} & \textbf{0.364} & 0.776 & 0.792 & 0.799 & \textbf{0.789} \\
 &    &  
   & \; \textsc{C}$\cdot$\textsc{R}        & \textbf{0.912} & \textbf{0.938} & \textbf{0.944} & 0.747 & \textbf{0.838} & 0.853 & \textbf{0.833} & \textbf{0.901} & \textbf{0.909} & \textbf{0.480} & 0.382 & 0.359 & \textbf{0.777} & 0.791 & \textbf{0.800} & \textbf{0.789} \\
&    &  
   & \; \textsc{C}$\cdot$\textsc{R}$\cdot$Qwen        & 0.901 & 0.934 & 0.940 & \textbf{0.749} & \textbf{0.838} & \textbf{0.854} & 0.831 & 0.899 & 0.906 & \textbf{0.480} & 0.382 & 0.359 & 0.775 & 0.790 & 0.798 & 0.788 \\

\midrule
\midrule

\multirow{7}{*}{\rotatebox{90}{DECODER}}
 & \multirow{3}{*}{Mean} 
   & \multirow{3}{*}{LLM2Vec}
   & \; Original & 0.938 & 0.964 & 0.967 & 0.835 & 0.904 & 0.913 & \textbf{0.854} & \textbf{0.922} & \textbf{0.929} & 0.440 & 0.408 & 0.358 & 0.814 & 0.827 & 0.824 & 0.822 \\
 &    &  
   & \; \textsc{C}$\cdot$\textsc{R}        & \textbf{0.941} & \textbf{0.965} & \textbf{0.968} & \textbf{0.839} & \textbf{0.907} & \textbf{0.916} & \textbf{0.854} & \textbf{0.922} & \textbf{0.929} & \textbf{0.500} & \textbf{0.424} & \textbf{0.372} & \textbf{0.826} & \textbf{0.831} & \textbf{0.827} & \textbf{0.828} \\
 &    &  
   & \; \textsc{C}$\cdot$\textsc{R}$\cdot$Qwen        & 0.940 & 0.964 & 0.967 & 0.834 & 0.904 & 0.912 & 0.853 & 0.921 & 0.928 & 0.480 & 0.421 & 0.366 & 0.821 & 0.829 & 0.825 & 0.825 \\
\cmidrule(l){2-20}
 & \multirow{3}{*}{Last} 
   & \multirow{3}{*}{Linq-Embed-Mistral}
   & \; Original & 0.944 & 0.967 & 0.969 & \textbf{0.800} & \textbf{0.885} & \textbf{0.895} & 0.876 & \textbf{0.937} & \textbf{0.942} & 0.460 & 0.407 & 0.360 & 0.810 & 0.828 & 0.830 & 0.823 \\
 &    &  
   & \; \textsc{C}$\cdot$\textsc{R}        & 0.942 & 0.967 & 0.969 & 0.798 & 0.882 & 0.892 & \textbf{0.877} & \textbf{0.937} & \textbf{0.942} & \textbf{0.500} & \textbf{0.423} & \textbf{0.373} & 0.815 & \textbf{0.830} & \textbf{0.832} & \textbf{0.826} \\
&    &  
   & \; \textsc{C}$\cdot$\textsc{R}$\cdot$Qwen        & \textbf{0.948} & \textbf{0.968} & \textbf{0.972} & 0.799 & \textbf{0.885} & \textbf{0.895} & 0.874 & 0.936 & 0.940 & \textbf{0.500} & \textbf{0.423} & \textbf{0.373} & \textbf{0.817} & \textbf{0.830} & \textbf{0.832} & \textbf{0.826} \\

\bottomrule
\end{tabular}
}
\caption{Performance of retrieval tasks with coreference resolution via Qwen2.5-7B-Instruct. The @k indicates the top k nDCG results. For each comparison, the higher score is highlighted in \textbf{bold}.}
\label{tab:qwen_retrieval}
\end{table*}

\section{Detailed Experimental Setup}
\subsection{Datasets}
\label{sec:exp_setup}
In processing the data for retrieval tasks, due to the substantial size of SQuAD2.0 and BoolQ datasets, we only use their validation data to construct the retrieval pool, as applying coreference resolution to the entire document set would be computationally intensive. For SQuAD2.0, we exclude all instances where answers are not available.

Among these datasets, BELEBELE, SQuAD2.0, and BoolQ, which contain answer information, are additionally utilized to evaluate the generation capabilities of our model. This allows us to demonstrate comprehensive effectiveness by assessing whether the model can generate improved responses to queries based on the retrieved documents.

\subsection{Prompt Templates}
\label{sec:exp_setup_prompt}
This section provides an overview of the prompt templates used in our experiments. 
\paragraph{Coreference Resolution}

Table~\ref{tab:prompt_cr} outlines the prompt applied for coreference resolution. This prompt instructs the model to act as a coreference resolution expert, replacing ambiguous pronouns with their explicit antecedents. The prompt includes examples demonstrating how pronouns should be resolved to their corresponding entities, ensuring consistent and accurate resolution.

\paragraph{QA inference}
For QA tasks, we utilize different prompts tailored to each dataset’s characteristics. Table~\ref{tab:prompt_boolq} shows the prompt for BoolQ, which presents the document and question in a straightforward format for yes/no answers. Table~\ref{tab:prompt_belebele} presents the prompt for Belebele, structured to handle multiple-choice questions with four options. Table~\ref{tab:prompt_squad} illustrates the prompt for SQuAD2.0, which explicitly instructs the model to provide concise answers to questions based on the given document.

\subsection{Hardware}
\label{sec:exp_setup_gpu}
We conducted our experiments using an Intel Xeon Gold 6230R @2.10GHz CPU, 376GB RAM, and an NVIDIA RTX A6000 48GB GPU. The software environment included nvidiadriver, CUDA, and PyTorch, running on Ubuntu 20.04.6 LTS.
\begin{table}[h!]
\centering
\scalebox{0.95}{
\small
\fbox{
\begin{minipage}{\dimexpr\linewidth-5\fboxsep-2\fboxrule\relax}
You are an expert in coreference resolution. Your task is to resolve all ambiguous pronouns and references in the provided document, replacing them with explicit and contextually accurate entities. Do not add any extra text or commentary—output only the fully resolved document.
\\ \\
Below are some examples:
\\ \\
Example 1: \\
Input: \\
Document: Alice, who was late, quickly ran to catch the bus because she missed her train. \\
Output: \\
Alice, who was late, quickly ran to catch the bus because Alice missed her train.
\\ \\
Example 2: \\
Input: \\
Document: Bob said he would finish his work today because he promised his manager. \\
Output: \\
Bob said that Bob would finish Bob's work today because Bob promised his manager.
\\ \\
Example 3: \\
Input: \\
Document: The committee stated that they would review the proposal after they received feedback. \\
Output: \\
The committee stated that the committee would review the proposal after the committee received feedback.
\\ \\
When you receive the input document (which always starts with "Document:"), please output only the resolved document text. \\ \\
Document: \{Document\}

\end{minipage}
}
}
\caption{Prompt template example for CR task.}
\label{tab:prompt_cr}
\end{table}

\begin{table*}[t] 
\centering
\small
\begin{tabular}{@{}l|llllllll@{}}
\toprule
\multirow{2}{*}{}  & \multicolumn{2}{c}{\textbf{Belebele}}                 & \multicolumn{2}{c}{\textbf{Bool Q}}                   & \multicolumn{2}{c}{\textbf{SQuAD v2.0}}               & \multicolumn{2}{c}{\textbf{NanoSCIDOCS}}              \\
                   & \multicolumn{1}{c}{original} & \multicolumn{1}{c}{CR} & \multicolumn{1}{c}{original} & \multicolumn{1}{c}{CR} & \multicolumn{1}{c}{original} & \multicolumn{1}{c}{CR} & \multicolumn{1}{c}{original} & \multicolumn{1}{c}{CR} \\ \midrule
Total words        & 44,258                       & 46,391                 & 320,991                      & 336,673                & 176,918                      & 184,348                & 354,405                      & 362,154                \\
AVG noun chunks    & 22.05                        & 22.73                  & 26.00                        & 26.70                  & 35.89                        & 36.75                  & 44.83                        & 44.81                  \\
AVG pronoun chunks & 2.70                         & 1.39                   & 2.36                         & 1.24                   & 2.85                         & 1.86                   & 4.39                         & 2.96                   \\ \bottomrule
\end{tabular}
\caption{Referential complexity computed using noun chunk detection in SpaCy~\citep{spacy2}. We observe that applying coreference resolution increases the number of noun chunks while reducing the number of pronoun chunks. This implies a reduction in referential ambiguity, thereby simplifying contextual understanding.\label{tab:referential_complexity}}
\end{table*}

\section{Coreferential Complexity}

\label{sec:referential_complexity}

Table~\ref{tab:referential_complexity} presents the number of noun and pronoun chunks before and after applying coreference resolution across different datasets. We define referential complexity as the degree of difficulty in understanding a given context, where a higher number of pronouns increases ambiguity in contextual comprehension.
The comparison between Table~\ref{tab:model_comparison} and Table~\ref{tab:referential_complexity} reveals that reduced referential complexity through coreference resolution correlates with improved retrieval performance, particularly in models using mean pooling strategies. When examining Table~\ref{tab:qa} and Table~\ref{tab:referential_complexity}, we observe that this reduction in referential complexity enhances QA performance across all model sizes, with smaller language models showing notable gains. These smaller models particularly benefit from the more explicit representation provided by coreference resolution, as demonstrated by their improved performance in tasks like BoolQ, Belebele, and SQuAD2.0.

\begin{table}[ht!]
\centering
\scalebox{0.95}{
\small
\fbox{
\begin{minipage}{\dimexpr\linewidth-5\fboxsep-2\fboxrule\relax}
Please refer to the given passage and choose the correct answer.\\ \\ 
P: \{Document\} \\ 
Q: \{Question\} \\ 
A: \{mc\_answer1\} \\ 
B: \{mc\_answer2\} \\
C: \{mc\_answer3\} \\ 
D: \{mc\_answer4\} \\ 
Answer:

\end{minipage}
}
}
\caption{Prompt template example for BELEBELE inference.}
\label{tab:prompt_belebele}
\end{table}
\begin{table}[ht!]
\centering
\scalebox{0.95}{
\small
\fbox{
\begin{minipage}{\dimexpr\linewidth-5\fboxsep-2\fboxrule\relax}
\{Document\} \\ \\
Question: \{Question\} \\ \\
Answer: 

\end{minipage}
}
}
\caption{Prompt template example for BoolQ inference.}
\label{tab:prompt_boolq}
\end{table}
\begin{table}[ht!]
\centering
\scalebox{1}{
\small
\fbox{
\begin{minipage}{\dimexpr\linewidth-5\fboxsep-2\fboxrule\relax}
\textbf{Instruction} \\
Please answer the question. \\ \\
\textbf{Conditions} \\
You must answer the question. with short answer. \\ \\
Document: \{Document\} \\ \\
Question: \{Question\} \\ \\
Answer:

\end{minipage}
}
}
\caption{Prompt template example for SQuAD2.0 inference.}
\label{tab:prompt_squad}
\end{table}

\clearpage

\onecolumn

\end{document}